\documentclass{article}

\usepackage[utf8]{inputenc}
\usepackage[margin=1.25in]{geometry}
\usepackage{authblk}
\usepackage{setspace}
\usepackage{graphicx}
\graphicspath{{./figures/}}
\usepackage{subcaption}
\usepackage{amsmath}
\usepackage{booktabs}
\usepackage{hyperref}
\usepackage{comment}
\usepackage{booktabs}

\usepackage[switch]{lineno}       
\usepackage[
  style=nejm, 
  citestyle=numeric-comp,
  sorting=none
]{biblatex}
\title{When Multi-Sensor Fusion Fails to Generalize: Cattle Posture Classification Under Animal-Level and Temporal Distribution Shift}

\author[*,1]{Leutrim Uka}
\author[*,2]{Severino Pinto}
\author[2]{Gundula Hoffmann}
\author[1,3]{Marina M.-C. Höhne}

\affil[1]{Institute of Computer Science, University of Potsdam, Potsdam, Germany.}
\affil[2]{Department of Sensors and Modelling, Leibniz Institute for Agricultural Engineering and Bioeconomy - ATB, Potsdam, Germany.}
\affil[3]{Department of Data Science in Bioeconomy, Leibniz Institute for Agricultural Engineering and Bioeconomy - ATB, Potsdam, Germany.}

\date{}

\begin{document}

\maketitle
\begingroup
\renewcommand\thefootnote{*}
\footnotetext{These authors contributed equally to this work.}
\endgroup
\begin{abstract} 
Automated cattle posture-classification systems frequently report near-perfect accuracy, yet their robustness under realistic deployment conditions remains largely unknown. In particular, it is unclear whether multimodal sensor fusion improves generalisation or leads models to rely on context-specific signals that fail under distribution shift. Here, we evaluate the robustness of automated posture classification (lying versus standing) using collar accelerometers, rumen-bolus sensors, and environmental measurements collected from a pasture-based beef cattle herd across two consecutive years (2024-2025). XGBoost served as the primary model, with Logistic Regression, Random Forest, and Long Short-Term Memory (LSTM) networks evaluated as comparative baselines. Model robustness was assessed under progressively more stringent evaluation protocols, ranging from conventional random train–test splits to leave-one-animal-out validation and cross-year evaluation on an independent cohort of previously unseen animals recorded one year later. While multimodal models achieved strong within-year performance (macro-F1 = 0.94), the performance declined substantially under cross-year evaluation (macro-F1 = 0.49). Unexpectedly, the collar-only model outperformed all multimodal configurations under temporal shift (macro-F1 = 0.54), indicating that additional physiological and environmental information did not improve robustness. Explainability analysis revealed persistent reliance on rumen-bolus activity and environmental variables even when predictive performance deteriorated, a pattern consistent with shortcut learning in which models exploit context-dependent signals that do not remain reliable under covariate shift. Distribution-shift diagnostics further confirmed substantial differences in the feature distributions between recording years. Our findings demonstrate that commonly used evaluation protocols can substantially overestimate real-world performance and that multimodal sensor fusion may reduce, rather than improve, robustness under temporal distribution shift. More broadly, the results highlight that benchmark accuracy alone is insufficient to assess deployment readiness and underscore the need for robustness-centred evaluation in livestock-monitoring research.
Together, our results highlight the importance of evaluating livestock-monitoring systems under realistic deployment conditions and demonstrate how explainability analysis can reveal hidden model dependencies and provide insight into model behaviour.
\end{abstract}
\vspace{1em}


\section{Introduction}
Monitoring animal behaviour is a key component of livestock management, as changes in activity patterns can provide early indicators of health and welfare problems. In particular, the balance between lying and standing behaviour in cattle has been widely associated with conditions such as lameness, environmental discomfort, and heat stress \cite{heinicke2018effects,Ito2010LyingCows, Tucker2008EffectSystem}. Continuous and automated monitoring of posture therefore has the potential to support early intervention and improve decision-making in modern farming systems.

Wearable accelerometers have become the dominant tool for automated cattle posture classification, enabling large-scale, continuous data collection in livestock environments \cite{Kleanthous2022DeepData}. Combined with machine learning methods, these systems allow the automatic classification of animal behaviour at fine temporal resolution. Reported posture classification performance frequently exceeds F1 scores of 0.9, creating the impression that automated cattle posture monitoring can already be performed reliably and robustly under practical condition \cite{Li2021DataClassification, Kleanthous2022DeepData, Bloch2023DevelopmentData}. 
Despite these promising results, it remains unclear whether the reported performance of current posture-classification models also reflects their robustness under real-world deployment conditions. A common limitation is the use of random or stratified train–test splits that allow observations from the same animals to appear in both training and test sets \cite{Robert2009EvaluationCattle, Li2022ClassificationMethods, Wu2022RecognisingCollar}. Such protocols can produce overly optimistic estimates of generalisation because models are evaluated on individuals whose behavioural patterns have already been observed during training. Even when subject-wise validation strategies, such as leave-one-animal-out (LOAO), are adopted \cite{Rahman2018CattleSensors, Kasfi2016ConvolutionalClassification}, evaluation is typically restricted to a single recording period. As a result, robustness under temporal distribution shift--a central challenge for practical deployment--remains largely unexplored.

In addition, most existing approaches rely on a single accelerometer stream, while multimodal extensions have largely remained within the same biomechanical domain\cite{Gonzalez2015BehavioralCattle, Umstatter2008AnSystems, Ungar2005InferenceCattle}. Modalities that capture physiological or environmental state remain largely unexplored in posture classification research, even though they may provide additional predictive information. Environmental conditions in particular have been shown to influence posture behaviour, with heat stress consistently associated with increased standing time and reduced lying duration in cattle \cite{heinicke2018effects,Herbut2018TheSystem}. Similarly, rumination behaviour is closely associated with resting periods, with cattle frequently ruminating while lying down \cite{Schirmann2012RuminationCows}. These biological relationships suggest that contextual information could improve posture classification beyond what can be inferred from movement signals alone.
However, contextual variables may also introduce a less obvious risk. Rather than learning posture-related behaviour, models may exploit correlations that are specific to a particular recording period, herd, or environmental context. Such dependencies can improve benchmark performance while failing to generalise when conditions change. This concern is particularly relevant for multimodal systems, where additional information may act as a proxy for recording-specific conditions rather than the target behaviour itself. Consequently, improvements in within-dataset accuracy do not necessarily imply improved robustness.
In this study, we investigate whether multimodal sensor fusion improves the robustness of cattle posture classification. Using a multimodal dataset collected from a pasture-based suckler cow herd across two consecutive years, we combine collar accelerometer data with physiological measurements from rumen boluses and environmental data from a weather station. Rather than relying on a single evaluation protocol, we assess model performance under increasingly stringent conditions, ranging from conventional random train-test splits to grouped animal-level evaluation, leave-one-animal-out evaluation, and cross-year evaluation on an independent cohort recorded one year later.
To understand not only how models perform but also why they perform, we additionally apply explainable artificial intelligence (XAI) to identify the features driving model predictions, determine whether multimodal models rely on meaningful signals or context-dependent proxies, and investigate how these dependencies change under temporal distribution shift.

The main contributions of this work are as follows:
\begin{itemize}
\item We demonstrate that commonly used evaluation protocols can substantially overestimate posture-classification performance, with predictive accuracy decreasing significantly under animal-level and temporal generalisation.
\item We show that multimodal sensor fusion, despite improving within-dataset performance, can reduce robustness under cross-year evaluation.
\item By using explainable AI for the distribution-shift analysis, we identify context-wise feature dependencies that remain influential despite decreased predictive performance, a pattern consistent with shortcut learning, whereby models rely on predictive but non-robust signals under distribution shift.
\end{itemize}

Together, these findings demonstrate that benchmark accuracy alone is insufficient to assess the model robustness under real-world deployment and highlight the need for robustness-centred evaluation in livestock-monitoring research.

\section{Related Work}
Cattle posture classification has been approached with a range of machine learning methods. Early work predominantly relied on classical statistical and machine learning methods applied to tri-axial accelerometer data from neck-, ear-, and limb-mounted sensors. Using Support Vector Machines, Linear Discriminant Analysis, and tree-based ensemble models, these studies established that lying and standing behaviour can be distinguished with reasonable reliability, with class sensitivities commonly reported at approximately 80\%\cite{Martiskainen2009CowMachines, Robert2009EvaluationCattle, VazquezDiosdado2015ClassificationSystem}. 

More recently, deep learning architectures have improved the performance further. Convolutional Neural Networks (CNNs) enable local pattern recognition from sensor signals, while recurrent models such as Long Short-Term Memory (LSTM) networks and Gated Recurrent Units (GRUs) capture temporal dependencies, with several studies reporting F1 scores exceeding 0.9 and in some cases reaching 0.99 for specific postures \cite{Bloch2023DevelopmentData, Wu2022RecognisingCollar, Kleanthous2022DeepData}. Transfer learning has also been explored as a strategy to improve performance under limited labelled data \cite{Kleanthous2022DeepData}. 

Despite these methodological advances, evaluation practices have remained comparatively largely unchanged, often failing to reflect realistic deployment conditions. Under cross-dataset evaluation, performance drops from approximately 94\% to 60\% in one direct comparison \cite{Bloch2023DevelopmentData}, highlighting limited generalisation despite increased model capacity.

A fundamental issue underlying many reported results is subject-level data leakage: numerous studies apply random or stratified splits without enforcing animal-level separation, allowing observations from the same individuals to appear in both training and test sets \cite{Robert2009EvaluationCattle, Li2022ClassificationMethods, Wu2022RecognisingCollar}. In this work \cite{Rahman2018CattleSensors}, the authors directly quantified the consequences, observing a decrease in the F1-score of more then 50 percentage points when switching from stratified to leave-one-animal-out (LOAO), also referred to as leave-one-group-out (LOGO), evaluation, with some behaviour classes falling from 0.91 to 0.15. These results demonstrate that models evaluated under leaky splits can appear highly accurate while failing on new individuals. Even when LOAO is adopted, robustness under temporal distribution shift remains almost entirely unexplored: most studies collect data over a narrow timeframe and do not assess whether models remain reliable as environmental conditions or herd composition change, which is a critical gap for real-world deployment \cite{Bloch2023DevelopmentData}.

In addition to soft evaluation practices, most posture classification systems rely on a single accelerometer signal, and multimodal extensions have largely remained within the same biomechanical domain, combining movement-derived signals such as GPS speed or orientation features \cite{Gonzalez2015BehavioralCattle, Ungar2005InferenceCattle}. Genuinely complementary modalities have received little attention despite clear biological motivation: heat stress increases standing time \cite{Herbut2018TheSystem} and rumination is closely coupled with lying behaviour \cite{Schirmann2012RuminationCows}, suggesting that environmental and physiological signals could support to disambiguate low-activity states that are difficult to distinguish from movement patterns alone. Whether such multimodal fusion improves generalisation or instead encourages models to exploit unstable context-specific correlations remains an open question. Addressing this question requires evaluation protocols that explicitly assess robustness under both cross-animal and cross-temporal distribution shift.

\section{Materials and Methods}
\subsection{Experimental Setup and Data Collection}
The study was reviewed and approved by the competent authority for licensing and notification procedures for animal experiments (LAVG) in Brandenburg, Germany (V6-2347/0-2022-11). It was conducted at a regenerative and ecological farm in eastern Brandenburg, Germany, on a pasture-based beef cattle herd under a rotational grazing system. Data were collected across two independent recording periods: a primary dataset in 2024 used for model development and an independent dataset in 2025 used for temporal validation. Ground-truth posture labels (lying, standing) were obtained through direct visual observation by trained personnel at a temporal resolution of one minute. Detailed information on the herd composition is provided in the Appendix. The data and source code used in this study are publicly available on GitHub\footnote{\url{https://github.com/leutrim-uka/cattle-posture-classification}} and Zenodo\footnote{\url{https://zenodo.org/records/19684521}}.

\subsection{Dataset and Sensor Modalities}
The primary dataset (2024) comprises 106 animals monitored on nine non-consecutive observation days between July and September, resulting in 24.231 minute-level labelled observations with matched sensor readings. The temporal validation dataset (2025) was collected under the same protocol one year later, comprising 95 animals and 13.949 labelled observations. Out of these cows, 43 are exclusive within the 2025 dataset, allowing us to test both generalisation to unseen cows and performance under time shift. 

Posture is labelled as \textit{lying} or \textit{standing}, where the latter includes idle standing, feeding, ruminating, and walking. The dataset is imbalanced at approximately 80\% standing and 20\% lying overall. Additionally, there is substantial variation across individuals; several animals did not exhibit both postures during the recording sessions.

The following three complementary sensing modalities were used, all synchronised to the one-minute resolution of the posture labels:

\textbf{Movement (collar-derived).} Each animal was equipped with an eShepheed\texttrademark{} neck collar (Gallagher Group Limited, Hamilton, NZ), which records inertial measurement unit (IMU) tick counts aggregated at six acceleration thresholds (40–240 mG) at one-minute resolution. Moreover, minute-to-minute changes in walking distance are derived from GPS coordinates.

\textbf{Physiological.} A subset of 72 animals from the 2024 dataset and all 43 cows from the 2025 dataset were additionally equipped with smaXtec® rumen boluses (smaXtec animal care GmbH, Graz, Austria), recording internal temperature and rumination signals at ten-minute intervals. 

\textbf{Environmental.} A nearby weather station recorded air temperature, relative humidity, wind speed, solar radiation, and ground temperature at fifteen-minute intervals, providing environmental context applicable uniformly across all animals.

\subsection{Data Preprocessing}
As part of the preprocessing, we first screened the collar-derived data for invalid measurements. Observations in which all IMU tick counters were simultaneously zero were removed, since these patterns likely reflected sensor malfunction or transmission failure rather than biologically plausible inactivity, affecting approximately 2.8\% of the dataset. In addition, the cumulative GPS odometer was transformed into minute-to-minute differences to eliminate spurious correlations with elapsed recording time and preserve only behaviourally meaningful movement information.

Rumen bolus features exhibiting near-zero variance across the dataset (e.g., calving index, water intake) were excluded prior to modelling, as features with minimal variability provide limited discriminative value and can introduce unnecessary noise into the learning process. Because the three sensor streams operate at different temporal resolutions, all data streams were temporally aligned using the collar data as the one-minute reference timeline. A left join strategy was applied to retain all collar observations, while lower-frequency modalities were forward-filled to the next available reading, corresponding to a maximum propagation window of ten minutes for rumen bolus data and fifteen minutes for weather observations. This filling procedure was based on the assumption that physiological and environmental processes evolve gradually over these time scales and therefore remain approximately stable between consecutive measurements.

\subsection{Modeling Approach}

Posture classification was formulated as a supervised binary classification task distinguishing lying from standing behaviour. XGBoost served as the primary model due to its strong performance on structured tabular data, computational efficiency, robustness to heterogeneous feature types, and compatibility with SHAP-based explainability methods. Logistic regression, random forest, and an LSTM network were evaluated as comparative baselines, where the LSTM was included to assess whether explicit sequential modelling provides additional benefit over conventional machine-learning approaches. As all models achieved highly comparable performance, XGBoost was selected for the subsequent analyses. Detailed results for all baseline models are provided in Appendix~\ref{app:model-comparison}.
Hyperparameters were tuned using five-fold grouped cross-validation within the training data, ensuring that no animal appeared in both training and validation folds.

To assess the impact of multimodal sensing, we train the models on features from different sensor combinations:
\begin{itemize}
\item \textbf{Collar-only:} Movement-related features derived from accelerometer and distance measurements.
\item  \textbf{Collar + Bolus/Weather:} Collar features combined with either physiological signals (rumen bolus) or environmental data (weather).
\item \textbf{All modalities:} Combination of movement, physiological, and environmental features.
\end{itemize}
This design enables a controlled comparison between biomechanical signals directly related to posture and contextual features that may improve performance but are potentially sensitive to distributional changes.

\subsection{Robustness Evaluation Under Distribution Shift}
Model robustness was evaluated under four complementary protocols of progressively increasing generalisation difficulty, ranging from conventional random splitting to strict temporal transfer. Together, these settings assess the impact of observation-level leakage, animal-level generalisation, and temporal distribution shift on classification performance.

\textbf{Random observation-level splits.} Following common practice in the literature \cite{Li2022ClassificationMethods, Peng2019ClassificationUnits}, the 2024 dataset was randomly partitioned into training and test sets at the observation level. Under this protocol, samples originating from the same animal may appear in both sets, allowing substantial subject overlap between training and evaluation data. Performance was estimated using five-fold cross-validation and aggregated across all observations. While this setting provides a useful reference point and facilitates comparison with previous studies, it does not assess generalisation to unseen animals.

\textbf{Grouped stratified splits.} The 2024 dataset was divided into 80\% training and 20\% test sets at the animal level, with stratification based on quantile-binned individual lying ratios to ensure comparable class distributions across splits despite heavy imbalance. In this setting, performance metrics were computed at the observation level by pooling all test samples, such that each observation contributes equally regardless of the individual animal identity. To reduce sensitivity to a particular partitioning, the entire procedure was repeated ten times with different random seeds, where results are reported as mean ± standard deviation across runs.

\textbf{Leave-one-animal-out (LOAO).} Models were trained on all but one animal and tested on the held-out individual, repeated for every animal in the dataset. This protocol eliminates animal-level leakage entirely and provides the strictest within-year measure of generalisation to unseen individuals. Performance metrics were computed per animal and then averaged across all animals.

\textbf{Cross-year temporal transfer.} Models trained on the full 2024 dataset were evaluated on the independent 2025 test set comprising only of the 43 previously unseen animals. This protocol introduces simultaneous animal-level and temporal distribution shift. To ensure comparability with the LOAO setting, performance metrics were computed per animal and then averaged across all animals.

The performance was quantified using macro-averaged F1 score as the primary metric to account for class imbalance and asymmetric error costs between the two posture classes:
\[
\text{Macro-F1} = \frac{1}{2} (F1_{\text{lying}} + F1_{\text{standing}}),
\]
where the class-wise F1 score is defined as the harmonic mean of precision and recall:
\[
F1 = 2 \cdot \frac{\text{Precision} \cdot \text{Recall}}{\text{Precision} + \text{Recall}}.
\]

Balanced accuracy, defined as the average recall across classes, was reported as a complementary metric:
\[
\text{Balanced Accuracy} = \frac{1}{2} (\text{Recall}_{\text{lying}} + \text{Recall}_{\text{standing}}).
\]

Macro-averaging ensures equal weighting across both posture classes despite the substantial imbalance between lying and standing observations. Importantly, the evaluation protocols differ not only in their degree of distribution shift, but also in the level at which performance is aggregated: grouped stratified evaluation reports observation-level performance, whereas LOAO and cross-year transfer assess robustness at the level of individual animals. 

\subsection{Explainability and Distribution Shift Diagnostics}
To investigate which features drive model predictions and whether these contributions remain stable across evaluation settings, SHAP (SHapley Additive exPlanations) \cite{lundberg2017unified} values were computed for the trained XGBoost models under both the within-year and cross-year configurations. This allows us to assess whether models rely on generalisable posture-related signals or on context-specific correlations tied to the recording conditions of the training data.

To characterise distributional differences between recording years, Principal Component Analysis (PCA) was fitted on standardised 2024 training features and and subsequently applied to both datasets in order to visualise covariate shift in the feature space. Additionally a logistic regression domain classifier was trained to discriminate between 2024 and 2025 samples using only the input features, with the resulting ROC-AUC score serving as a quantitative measure of domain separability and cross-year distribution mismatch. Finally, to isolate the contribution of distribution shift from changes in task difficulty, a LOAO evaluation was also conducted exclusively within the 2025 dataset, allowing a direct comparison of intrinsic class separability between years independent of cross-year transfer.

\section{Results and Discussion}
In the following, we present the results across the three different evaluation protocols and various feature configurations. Unless stated otherwise, results are reported on the subset of animals equipped with both collar and rumen bolus sensors (N=72) to enable fair comparison across all sensing modalities. Full-cohort results (N=106) based on the collar differ only marginally from the bolus subset. For completeness, they are provided in the Appendix.

\subsection{The Illusion of Performance Under Standard Evaluation}
\begin{figure}[h]
    \centering
    \includegraphics[width=1\linewidth]{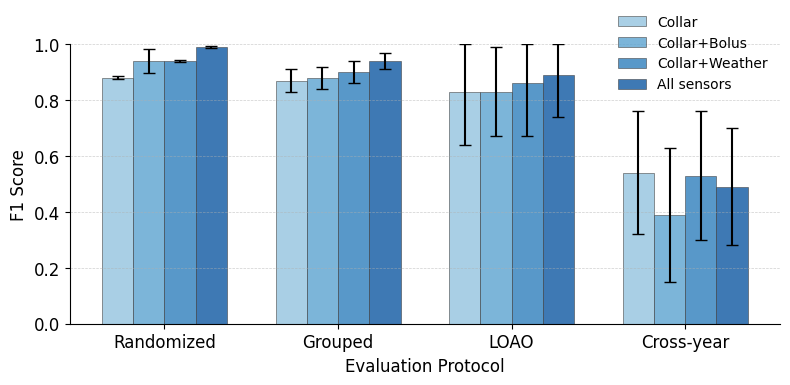}
    \caption{Posture-classification performance across evaluation protocols of increasing generalisation difficulty. Bars show macro-F1 scores for different sensor configurations, and error bars indicate the corresponding standard deviation. The XGBoost performance is highest under random splitting and remains strong under within-year grouped evaluation. However, both accuracy and stability decrease under leave-one-animal-out (LOAO) evaluation, revealing substantial variability in generalisation across individual animals. The largest decline occurs under cross-year transfer, where models trained on 2024 data are evaluated on an independent cohort recorded in 2025. Notably, multimodal models provide the strongest performance under within-year evaluation, whereas the collar-only model achieves the highest performance under temporal transfer, highlighting the limited robustness of contextual modalities under distribution shift. The exact values are shown in Table ~\ref{tab:main_results}}
    \label{fig:tenrun_vs_logo}
\end{figure}

We compare posture-classification performance across evaluation protocols of increasing stringency, ranging from conventional random observation-level splits to strict cross-year temporal transfer (Figure~\ref{fig:tenrun_vs_logo}, Table~\ref{tab:main_results}). 

Under random observation-level splitting, where samples from the same animals can appear in both training and test sets, classification performance approaches near-perfect discrimination across all sensor configurations, with the multimodal model achieving a macro-F1 of 0.99 (Table~\ref{tab:main_results}).

Performance remains high under grouped animal-level splits, where all observations from an individual animal are assigned exclusively to either training or test data. In this setting, the multimodal model achieves a macro-F1 of 0.94, with consistent improvements observed as additional sensing modalities are incorporated. Relative to the collar-only baseline (macro-F1 = 0.87), weather and bolus information appear to provide substantial predictive value.
However, this apparent robustness deteriorates under increasingly realistic evaluation conditions. Under LOAO evaluation, the best-performing multimodal configuration reaches a macro-F1 of 0.89. While average performance remains comparatively high, the standard deviation of the lying-class F1 score is 0.27, indicating that aggregate metrics conceal substantial failures for individual animals.

\begin{table*}[h]
\centering
\caption{Posture-classification performance of the XGBoost model across evaluation settings of increasing stringency. Results are reported as mean $\pm$ standard deviation.}
\label{tab:main_results}
\begin{tabular}{llcccc}
\hline
Setting &
Feature set &
Macro-F1 &
F1 (Standing) &
F1 (Lying) &
Balanced Acc \\
\hline

Random &
Collar &
$0.88 \pm 0.005$ &
$0.95 \pm 0.002$ &
$0.80 \pm 0.009$ &
$0.88 \pm 0.005$ \\
 &
Collar + Weather &
$0.94 \pm 0.003$ &
$0.98 \pm 0.001$ &
$0.91 \pm 0.005$ &
$0.95 \pm 0.004$ \\
 &
Collar + Bolus &
$0.94 \pm 0.044$ &
$0.98 \pm 0.002$ &
$0.91 \pm 0.069$ &
$0.94 \pm 0.044$ \\
 &
All sensors &
$\mathbf{0.99 \pm 0.002}$ &
$\mathbf{0.99 \pm 0.000}$ &
$\mathbf{0.99 \pm 0.005}$ &
$\mathbf{0.99 \pm 0.003}$ \\
\hline

Grouped &
Collar &
$0.87 \pm 0.04$ &
$0.95 \pm 0.02$ &
$0.79 \pm 0.07$ &
$0.87 \pm 0.05$ \\
 &
Collar + Bolus &
$0.88 \pm 0.04$ &
$0.96 \pm 0.02$ &
$0.81 \pm 0.07$ &
$0.88 \pm 0.05$ \\
 &
Collar + Weather &
$0.90 \pm 0.04$ &
$0.96 \pm 0.02$ &
$0.85 \pm 0.06$ &
$0.91 \pm 0.05$ \\
 &
All sensors &
$\mathbf{0.94 \pm 0.03}$ &
$\mathbf{0.97 \pm 0.01}$ &
$\mathbf{0.90 \pm 0.05}$ &
$\mathbf{0.94 \pm 0.04}$ \\
\hline

LOAO &
Collar &
$0.83 \pm 0.19$ &
$0.94 \pm 0.12$ &
$0.73 \pm 0.31$ &
$0.89 \pm 0.14$ \\
 &
Collar + Bolus &
$0.83 \pm 0.16$ &
$0.95 \pm 0.05$ &
$0.70 \pm 0.30$ &
$0.87 \pm 0.14$ \\
 &
Collar + Weather &
$0.86 \pm 0.19$ &
$0.95 \pm 0.12$ &
$0.77 \pm 0.32$ &
$0.90 \pm 0.16$ \\
 &
All sensors &
$\mathbf{0.89 \pm 0.15}$ &
$\mathbf{0.96 \pm 0.12}$ &
$\mathbf{0.81 \pm 0.27}$ &
$\mathbf{0.93 \pm 0.11}$ \\
\hline

Cross-year &
Collar &
$\mathbf{0.54 \pm 0.22}$ &
$\mathbf{0.87 \pm 0.13}$ &
$\mathbf{0.52 \pm 0.21}$ &
$\mathbf{0.67 \pm 0.24}$ \\
 &
Collar + Weather &
$0.53 \pm 0.23$ &
$0.84 \pm 0.20$ &
$0.49 \pm 0.23$ &
$0.65 \pm 0.26$ \\
 &
Collar + Bolus &
$0.39 \pm 0.24$ &
$0.86 \pm 0.17$ &
$0.14 \pm 0.18$ &
$0.52 \pm 0.32$ \\
 &
All sensors &
$0.49 \pm 0.21$ &
$0.86 \pm 0.17$ &
$0.41 \pm 0.24$ &
$0.62 \pm 0.25$ \\
\hline

\end{tabular}
\end{table*}
The discrepancy becomes most evident under cross-year evaluation. When models trained on 2024 data are applied to an independent cohort recorded one year later, macro-F1 decreases from 0.94 to 0.49 for the multimodal model, representing a relative reduction of almost 50\%. Notably, the collar-only model becomes the best-performing configuration under temporal transfer, outperforming all multimodal variants. This reversal contrasts sharply with the within-year results and suggests that the apparent benefit of contextual information does not translate to changing deployment conditions.

Taken together, these results demonstrate that performance estimates obtained under commonly used evaluation protocols can substantially overstate real-world generalisation capability. Models that achieve a high accuracy under random or within-year evaluation may exhibit a significantly lower performance when tested on previously unseen animals and temporally shifted data.

\subsection{Individual Variability Reveals Hidden Failure Modes}
The high standard deviations shown in Figure~\ref{fig:tenrun_vs_logo} already indicate that performance under LOAO evaluation is substantially more variable than within-year grouped evaluation. Most notably, the standard deviation of the lying-class F1 score increases from 0.05 under grouped splits to more than 0.27 under LOAO, revealing pronounced heterogeneity in model generalisation across individual animals.

\begin{figure}[h!]
    \centering
\includegraphics[width=0.8\linewidth]{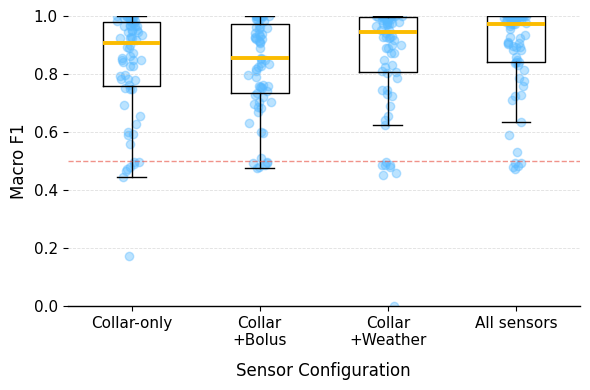}
    \caption{Visualization of the per-animal F1 score distribution under leave-one-animal-out (LOAO) evaluation, illustrating substantial variability in model performance across individuals. The distribution is highly skewed with a pronounced lower tail, indicating that while average performance remains relatively high, a subset of animals exhibits near chance-level classification. This highlights hidden failure modes that are not captured by aggregated metrics and underscores the importance of evaluating robustness at the individual-animal level.}
    \label{fig:placeholder}
\end{figure}

To better understand the source of this variability, we examine the distribution of per-animal performance scores shown in Figure~\ref{fig:placeholder}. Rather than exhibiting a compact and approximately symmetric spread around the mean, the per-animal F1 scores reveal a strongly skewed distribution with a pronounced lower tail containing several severely underperforming individuals, some approaching near-chance classification performance. Importantly, the distribution is not driven by a small number of isolated outliers, but by a systematic subgroup of animals for which the model generalises poorly despite strong aggregate benchmark scores.

The lower variance in grouped evaluation is partly due to structural factors: averaging over multiple animals per split and repeating across ten runs naturally smooth out individual extremes, resulting in a summary score that reflects herd-level tendencies rather than the reliability of individual animals. Therefore, standard evaluation protocols can give a misleading impression of robustness. LOAO, on the other hand, evaluates each animal individually, such that the resulting standard deviation directly reflects variability across individuals rather than across random partitions. Although the macro-F1 score is not designed to express this type of variance, it is precisely this limitation that makes reporting standard deviation an important addition to the LOAO evaluation score.

\subsection{Multimodal Models Improve Benchmark Performance but Fail Under Distribution Shift}

\begin{figure}[h]
    \centering
    \includegraphics[width=1\linewidth]{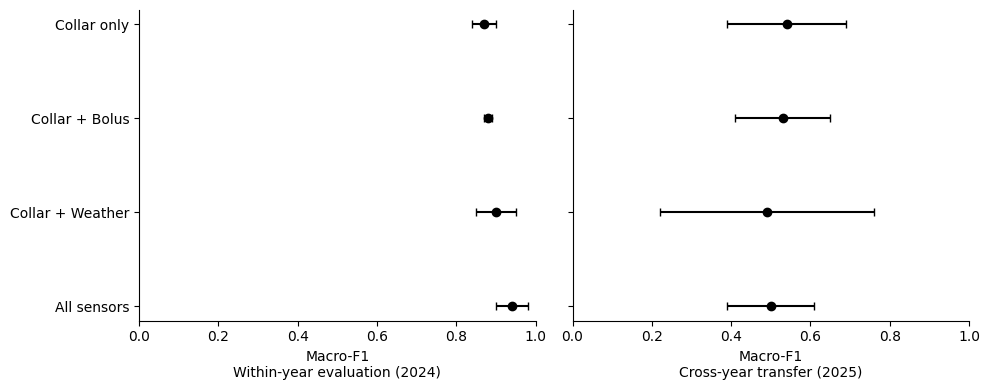}
    \caption{Effect of multimodal sensor fusion on posture-classification performance under within-year evaluation and cross-year transfer. Points indicate macro-F1 scores and error bars represent standard deviation. Under within-year evaluation on the 2024 dataset (left), performance increases consistently as additional physiological and environmental modalities are incorporated, with the full multimodal configuration achieving the highest accuracy. In contrast, under cross-year transfer to an independent 2025 cohort (right), this ranking reverses: multimodal models exhibit substantially larger performance degradation than the collar-only baseline and no longer provide a performance advantage. The strongest decline is observed for models incorporating environmental features, suggesting that contextual information introduces dependencies that do not generalise across recording periods. These results demonstrate that improvements in benchmark performance do not necessarily translate into improved robustness under temporal distribution shift.}
    \label{fig:fig3}
\end{figure}

Beyond evaluation protocol and individual-level variability, model performance is strongly influenced by input modality. Within the 2024 dataset, every additional modality improves macro-F1 relative to the collar-only baseline, with the combination of all sensors yielding the largest gain (Table~\ref{tab:main_results}). Weather features drive the most consistent improvement, particularly for the lying class, while rumen bolus features show limited independent contribution and improve performance mainly in combination with environmental variables (Figure~\ref{fig:fig3}, left).

However, under cross-year transfer, this pattern reverses. The collar-only model achieves a macro-F1 of 0.54 on the 2025 test set, while the multimodal model drops to 0.49. 

While F1 scores exceeding 0.9 have been reported for posture classification under within-year conditions \cite{Bloch2023DevelopmentData, Wu2022RecognisingCollar, Kleanthous2022DeepData}, these estimates are obtained under evaluation protocols that do not reflect temporal distribution shifts. A single direct cross-dataset comparison available in the literature reports a drop from approximately 94\% to 60\% accuracy \cite{Bloch2023DevelopmentData}, consistent with the magnitude of degradation observed here, and suggesting that temporal fragility may be a systematic property of current models rather than an artefact of this study.

Prior multimodal extensions in this domain have largely combined signals within the same biomechanical domain \cite{Gonzalez2015BehavioralCattle, Ungar2005InferenceCattle}. The biological motivation for incorporating environmental and physiological signals is clear: heat stress influences standing duration \cite{Herbut2018TheSystem} and rumination is tightly coupled with lying behaviour \cite{Schirmann2012RuminationCows}, but our results demonstrate that including these signals without architectural constraints introduces context-specific dependencies that reduce rather than improve cross-year robustness.

\subsection{Explainable AI Reveals Shortcut Learning in Multimodal Models}
To investigate why multimodal models fail under temporal shift, we analysed feature attribution patterns using SHAP  (SHapley Additive exPlanations) \cite{lundberg2017unified}.

\begin{figure}[h!]
    \centering
    \includegraphics[width=0.8\linewidth]{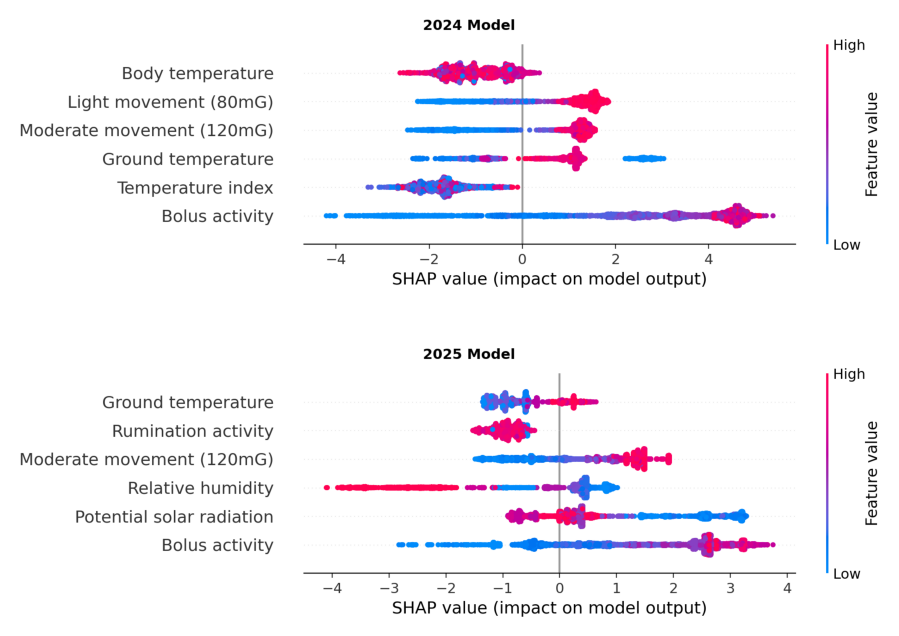}
    \caption{SHAP summary plots for the multimodal model under within-year (2024, top) and cross-year (2025, bottom) evaluation. Features are ordered by their average contribution to model predictions, with each point representing a single observation and colour indicating the underlying feature value. A notable finding is the persistent dominance of bolus activity across both years despite the substantial deterioration in cross-year classification performance. Together with the performance results, this suggests that the model continues to rely on feature relationships that are highly predictive within the training domain but less reliable under temporal distribution shift.}
    \label{fig:shap}
\end{figure}

In the SHAP summary plots, shown in Figure~\ref{fig:shap} for 2024 on the top and 2025 on the bottom, features are ranked by their overall contribution to model predictions, with higher-ranked features exerting greater influence on the predicted posture class. Each point represents a single observation, where the horizontal position indicates the magnitude and direction of the feature's contribution and the colour encodes the corresponding feature value.

From the resulting SHAP values, we can observe that bolus activity remains the most influential feature in both settings. Under within-year evaluation, this reliance is associated with strong predictive performance (macro-F1 = 0.94), indicating that bolus activity is highly informative within the 2024 recording context. However, when the same model is evaluated on the independent 2025 cohort, overall performance decreases substantially (macro-F1 $\approx$ 0.50), while bolus activity retains its dominant position in the SHAP ranking. The model therefore continues to rely heavily on the same feature despite a strong decrease in predictive performance under cross-year evaluation.

The cross-year performance breakdown across sensor configurations provides further insight into this behaviour (Table~\ref{tab:main_results}). Notably, the collar-only model achieves the highest cross-year performance (macro-F1 = 0.54), outperforming all multimodal variants. The addition of bolus features reduces the lying-class F1 score from 0.52 to 0.14, indicating that the contextual signal does not merely fail to contribute useful information but actively interferes with generalisation. Environmental variables exhibit a similar, albeit less pronounced, pattern.

Taken together, these observations suggest that the model has learned context-dependent feature–label relationships that are highly predictive within the 2024 recording environment but fail to remain reliable under temporal distribution shift. This pattern is consistent with shortcut learning \cite{Geirhos2020ShortcutNetworks} and Clever-Hans effects, in which models exploit predictive but non-generalising regularities in the training data. Such failure modes are precisely the type of hidden model behaviour that explainability methods are designed to expose \cite{lapuschkin2019unmasking, bykov2023dora, bykov2023mark}.

\subsection{Temporal Distribution Shift Between Years Explains Performance Degradation}
With the SHAP analysis we identified features the model relies on and why those dependencies fail under cross-year evaluation. To visualize the magnitude and structure of the underlying distributional change, we fit a PCA model on the standardised 2024 training data and project both years into the resulting low-dimensional feature space, i.e., onto the first two principal components, which cover approximately 58\% of the data variance. 
\begin{figure}[h!]
    \centering
    \includegraphics[width=0.6\linewidth]{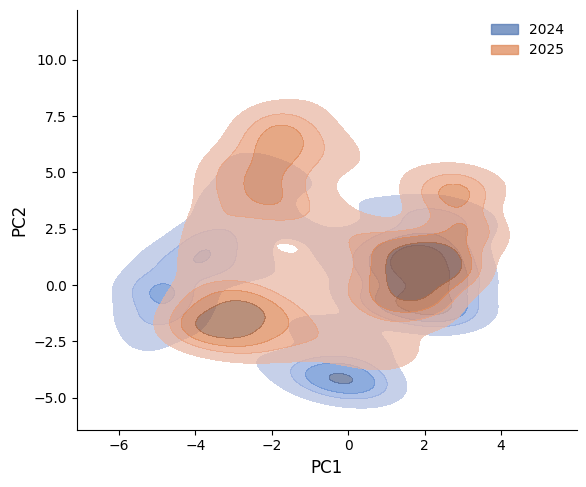}
    \caption{ Visualization of the distribution shift between recording years.
The feature space for the 2024 (training) and 2025 (test) datasets is plotted based on the first two PCA components, showing a substantial shift in the joint distribution of the input features. A domain classifier further confirms that the two datasets are statistically distinguishable. This distributional mismatch contributes to the observed degradation in model performance under cross-year evaluation.}
    \label{fig:distshift}
\end{figure}
As shown in Figure~\ref{fig:distshift}, samples from 2024 and 2025 occupy systematically different regions of the projected feature space, indicating a substantial shift in the joint feature distribution between recording years.

To quantify this separation, we train a domain classifier to distinguish between samples from the two years. When trained on the multimodal features, the classifier achieves a high ROC-AUC score of 0.99, confirming that the two datasets are statistically distinguishable and that a significant covariate shift is present. Restricting the input of the model to movement-only features reduces the ROC-AUC score to 0.69. Taken together, these results indicate that performance degradation under cross-year evaluation arises from two interacting factors: the model's reliance on context-specific proxy signals, as identified through SHAP analysis, and a broader mismatch in the underlying input distribution that affects both contextual and movement-derived features.

\begin{figure}[h]
    \centering
    \includegraphics[width=0.6\linewidth]{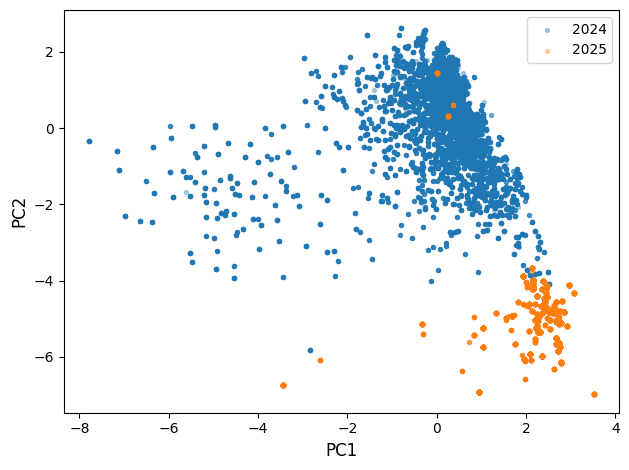}
    \caption{Visualization of the temporal distribution shift in collar-derived movement features between 2024 and 2025. The PCA model was fitted on standardised 2024 training data and applied to both years. Each point represents a minute-level observation coloured by the corresponding recording year. Despite excluding environmental and physiological variables, observations from 2024 and 2025 remain clearly separated, indicating that temporal distribution shift also affects the movement signals that are used for posture classification.
    }
    \label{fig:bolus-pca}
\end{figure}

The PCA projection in Figure~\ref{fig:distshift} demonstrates a substantial shift in the joint feature distribution between recording years. However, because the multimodal feature space includes environmental and physiological variables, it remains unclear whether this separation is driven primarily by contextual signals. To address this question, we repeated the analysis using collar-derived movement features only (Figure~\ref{fig:bolus-pca}). Despite excluding all environmental and physiological information, observations from 2024 and 2025 remain clearly separated, indicating that temporal distribution shift extends beyond contextual variables and also affects the movement signals underlying posture classification.

\section{Limitations}
Several limitations should be considered when interpreting the results and their implications for real-world deployment. 

\textbf{Temporal coverage and environmental diversity.} The dataset spans a relatively small number of observation days within each year (2024: 9 days between July and September; 2025: 3 days in August), and both recording periods are restricted to summer months and pre-noon observation windows. This narrow temporal coverage amplifies correlations between contextual features and specific recording conditions, increasing the risk that models exploit proxy signals rather than generalisable posture-related patterns--and a dynamic directly observed in the SHAP analysis.

\textbf{Single herd and management context.} The study is conducted on a single herd under a specific pasture-based management system. While this reflects realistic conditions for beef cattle production, the findings may not directly transfer to other production systems, breeds, or housing environments. Differences in sensor placement protocols, herd behaviour, and local environmental variability could alter both feature distributions and model performance that are not captured by this study.

\textbf{Binary posture formulation.} Posture classification is treated as a binary task distinguishing lying from standing. The standing class, however, aggregates several distinct behavioural states such as feeding, walking, and ruminating, introducing within-class heterogeneity that reduces separability and may cause the model to rely on coarser signals. Reported performance metrics therefore likely underestimate the difficulty of more fine-grained behaviour recognition tasks, and the binary framing should be considered when comparing results across studies.

\textbf{Class imbalance.} The dataset exhibits a notable imbalance between lying and standing observations (80\% standing, 20\% lying), which can affect model performance. In addition to the observation-level class imbalance, the variability among individual cows poses another challenge. Several animals exhibited only one posture class during the recording period. The consistently lower F1 scores for the lying class across experimental conditions may be partly attributable to this imbalance, and results should be interpreted accordingly.

\textbf{Label quality and annotation noise.} Ground truth labels are derived from manual observation, which introduces a degree of uncertainty at class boundaries where transitional behaviours, such as the onset of lying or rising, are difficult to assign unambiguously. Borderline observations of this kind may act as outliers that reduce apparent class separability and contribute to the lower intrinsic separability observed in the 2025 data. A more granular annotation protocol, or explicit modelling of transitional states, could reduce this source of noise.

\textbf{Model scope.} Although multiple architectures were evaluated, the analysis centres on tree-based models, which were consistently the best-performing class across configurations. The observed trends in robustness and feature reliance may differ for other model families, such as recurrent or transformer-based architectures that can explicitly model temporal dependencies. 

\textbf{Correlation versus causation in XAI findings.} The SHAP analysis identifies which features contribute most strongly to model predictions and, when interpreted alongside the temporal transfer experiments, suggests that contextual variables may encode recording-specific information that does not generalise reliably across years. However, feature-attribution methods quantify predictive relevance rather than causal influence. Consequently, the observed feature dependencies should be interpreted as model-specific associations and not as evidence that these variables causally determine posture behaviour or directly drive the observed performance degradation. Establishing causal relationships would require either controlled interventions or dedicated causal inference frameworks, which were beyond the scope of the present study.

Despite these limitations, this study provides a systematic and realistic evaluation of model robustness under cross-year distribution shift, combining performance analysis, explainability methods, and distributional characterisation to identify failure modes that are invisible under standard within-year evaluation protocols.

\section{Conclusion}

In this study, we investigated the robustness of machine learning models for cattle posture classification using multimodal wearable sensor data. Although high predictive performance is commonly reported under standard evaluation protocols, our results demonstrate that these estimates can substantially overstate real-world reliability. While multimodal models achieved a macro-F1 of up to 0.99 under conventional random splits and 0.94 under within-year evaluation, performance declined to 0.49 when evaluated on an independent cohort recorded one year later. Notably, the collar-only model outperformed all multimodal configurations under temporal transfer, indicating that additional contextual information does not necessarily translate into improved robustness.
By combining increasingly stringent evaluation protocols with explainability analysis and distribution-shift diagnostics, we identify two key factors underlying the discrepancy between benchmark and deployment performance. First, standard evaluation protocols can provide overly optimistic estimates of generalisation by masking substantial variability across individual animals. Second, multimodal models rely on context-dependent feature–label relationships that are highly predictive within the training environment but fail to remain reliable under temporal distribution shift.

These findings have implications beyond cattle posture classification. In many livestock-monitoring applications, contextual variables such as environmental, physiological, or management-related signals are increasingly incorporated to improve predictive performance. Our results show that apparent gains obtained under benchmark conditions may instead reflect reliance on unstable proxies that fail under realistic deployment scenarios. Consequently, high classification accuracy alone should not be interpreted as evidence of robustness.

Future research should therefore move beyond within-dataset benchmarks and adopt evaluation frameworks that explicitly assess generalisation across animals, time periods, and management conditions. Equally important, explainability methods should be integrated as routine robustness diagnostics rather than used solely for model interpretation. More broadly, our findings highlight the need to shift from benchmark-centric evaluation towards robustness-centric evaluation in livestock-monitoring research.

\section*{Acknowledgments}
This research was funded by the German Federal Ministry of Agriculture, Food and Regional Identity (BMLEH) based on a decision of the Parliament of the Federal Republic of Germany, granted by the Federal Office for Agriculture and Food (BLE; ref. 28DE204A21 ). Moreover, the work was funded by the Federal Ministry of Research, Technology and Space through the projects DCropS4OneHealth (ref. 16LW0528K) and REFRAME (ref. 01IS24073B). 
\printbibliography

\clearpage
\appendix

\section{Appendix}

\subsection{Herd Composition}
The data for this study were sourced from a herd of 180 cows in Brandenburg, Germany. Since April 2024, 164 individuals have been equipped with the eShepherd collar. 106 animals from this herd are part of the 2024 dataset, observed on 9 different days between July and September. This herd includes steers, pregnant cows, and heifers. 

Among these animals, 78 are female and 28 male. The herd shows heterogeneous breed composition: 53 animals are Aberdeen Angus (AA), 12 Red Angus (RA), and 6 Salers (SR), while the remaining 35 animals are of mixed breed, specifically an Aberdeen Angus × Salers cross.

The age distribution is skewed towards younger animals, with half of the herd being three years old or younger. Two-year-old animals form the largest age group, while the oldest animal in the dataset is eight years old. Age was calculated from the recorded date of birth relative to the observation date.

In the 2025 dataset, 7 animals are Aberdeen Angus (AA), 1 is Red Angus (RA), and 35 are of mixed breed. Of these, 12 cows are female, 1 is male, while the information for the other 22 cows is missing.

\section{Comparison of Alternative Classification Models}
\label{app:model-comparison}

To assess whether the observed findings depend on the choice of classifier, we compared XGBoost against Logistic Regression, Random Forest, and a Long Short-Term Memory (LSTM) network under the grouped within-year evaluation protocol. The LSTM baseline was included to evaluate whether explicit sequential modelling provides an advantage over conventional machine-learning approaches when classifying cattle posture from sensor data.

Table~\ref{tab:model_comparison} shows that all model families achieved highly similar performance when trained on collar-derived features alone. Macro-F1 scores ranged from 0.87 to 0.88, with virtually identical performance for both posture classes. These results indicate that classifier selection had only a minor influence on predictive performance under the within-year evaluation setting. Consequently, XGBoost was selected as the primary model for the main analyses due to its strong performance, computational efficiency, and compatibility with SHAP-based explainability methods.

The LSTM results further show that the relative contribution of different sensing modalities is largely independent of model architecture. Similar to the tree-based models, environmental variables improved within-year classification performance, whereas bolus-derived physiological features provided limited additional benefit. This consistency suggests that the main findings reported in the manuscript are not specific to a particular modelling approach.

\begin{table}[h]
\centering
\caption{Comparison of alternative classification models under the grouped within-year evaluation protocol (2024). Results are reported as mean $\pm$ standard deviation across repeated grouped stratified splits.}
\label{tab:model_comparison}
\begin{tabular}{llcccc}
\hline
Model &
Feature Set &
Macro-F1 &
F1 (Standing) &
F1 (Lying) &
Balanced Accuracy \\
\hline

Logistic Regression &
Collar &
$0.88 \pm 0.05$ &
$0.95 \pm 0.02$ &
$0.80 \pm 0.07$ &
$0.87 \pm 0.04$ \\

Random Forest &
Collar &
$0.88 \pm 0.04$ &
$0.95 \pm 0.01$ &
$0.80 \pm 0.07$ &
$0.87 \pm 0.04$ \\

XGBoost &
Collar &
$0.88 \pm 0.04$ &
$0.95 \pm 0.01$ &
$0.80 \pm 0.07$ &
$0.87 \pm 0.04$ \\

\hline

LSTM &
Collar &
$0.87 \pm 0.05$ &
$0.95 \pm 0.02$ &
$0.80 \pm 0.08$ &
$0.89 \pm 0.05$ \\

LSTM &
Collar + Weather &
$0.93 \pm 0.03$ &
$0.97 \pm 0.01$ &
$0.89 \pm 0.04$ &
$0.96 \pm 0.01$ \\

LSTM &
Collar + Bolus &
$0.87 \pm 0.05$ &
$0.94 \pm 0.02$ &
$0.80 \pm 0.08$ &
$0.90 \pm 0.05$ \\

LSTM &
All sensors &
$0.91 \pm 0.03$ &
$0.96 \pm 0.02$ &
$0.86 \pm 0.05$ &
$0.94 \pm 0.03$ \\

\hline
\end{tabular}
\end{table}

\end{document}